\documentclass[preprint,12pt,authoryear]{elsarticle}

\usepackage{amssymb}
\usepackage{amsmath}
\PassOptionsToPackage{hyphens}{url}
\usepackage[colorlinks]{hyperref}
\usepackage{graphicx}
\newcommand{\absdiv}[1]{%
  \par\addvspace{.5\baselineskip}
  \noindent\textbf{#1}\quad\ignorespaces
}
\usepackage{natbib}

\journal{Journal of Biomedical Informatics}

\begin{document}
\begin{frontmatter}


\title{Leveraging Few-Shot Learning and Large Language Models for Analyzing Blood Pressure Variations Across Biological Sex from Scientific Literature}

\author[label1]{Yuting Guo\corref{cor1}} 
\ead{yuting.guo@emory.edu}
\cortext[cor1]{Corresponding author}
\author[label1]{Seyedeh Somayyeh Mousavi} 
\ead{seyedeh.somayyeh.mousavi@emory.edu}
\author[label1]{Yao Ge} 
\ead{yao.ge@emory.edu}
\author[label2]{Madhumita Baskaran} 
\ead{baskaranm9906@acom.edu}
\author[label1,label3]{Reza Sameni} 
\ead{rsameni@dbmi.emory.edu}
\author[label1,label3]{Abeed Sarker} 
\ead{abeed@dbmi.emory.edu}
\affiliation[label1]{organization={Emory University},
            city={Atlanta},
            postcode={30322}, 
            state={GA},
            country={US}}
            
\affiliation[label2]{organization={Alabama College of Osteopathic Medicine},
            city={Dothan},
            postcode={36303}, 
            state={AL},
            country={US}}

\affiliation[label3]{organization={Georgia Institute of Technology},
            city={Atlanta},
            postcode={30332}, 
            state={GA},
            country={US}}

\begin{abstract}
\absdiv{Background}
Current blood pressure (BP) technologies and standards were established decades ago, and these standards are still used worldwide today, often without adjusting BP readings for individual demographic factors such as sex and age. While these standards provide useful guidelines and help identify at-risk patients, they are not fully reliable for diagnosis due to the lack of demographic considerations. This study aims to assess the feasibility of using large language models (LLMs) for the automated extraction of BP-related information from the scientific literature, with a focus on biological sex-based distinctions in BP distributions.

\absdiv{Method}
We employed natural language processing (NLP) methods to extract the means and standard deviations of BP values from the literature, distinguishing by biological sex. We developed a Solr-based search engine to retrieve scientific articles containing BP-related keywords and biological sex indicators from PubMed. From the retrieved articles, we created a manually reviewed subset comprising 213 articles including 90 cases that reported BP values based on biological sex. We experimented with one few-shot learning method and two zero-shot LLM-based methods---LLaMA3 and GPT-3.5---to extract the mean and standard deviations of BP values, and the associated biological sex. Based on the automatically-extracted information, we generated heatmaps and contour plots to study the variations of BP values across biological sex. 

\absdiv{Results}
The inter-annotator agreement (IAA) between the two annotators measured using Cohen's kappa~\cite{McHugh2012} was 0.74. The best performing system was LLaMA3 with an F$_1$ score of 0.85 ($\pm$0.00). The few-shot learning method (DANN) exhibited low performance with an average F$_1$ score of 0.30 ($\pm$0.01). GPT-3.5 achieved moderate performance with an average F$_1$ score of 0.67 ($\pm$0.04). The countor plots show that males tend to exhibit higher BP values than females. 

\absdiv{Conclusions}
Our results demonstrate that LLMs can be reliable in extracting population-level BP and biological sex information from clinical literature. They also 
outperform traditional low-shot information extraction systems in this context, showcasing their ability to extract BP-related information more accurately and efficiently. By employing LLMs, we provide a scalable framework for analyzing demographic differences in BP and emphasize the broader utility of LLMs in addressing similar challenges in biomedical research.

\end{abstract}



\end{frontmatter}

\section{Introduction}
Blood pressure (BP) monitoring is critical in diagnosing and mitigating the risk of cardiovascular diseases (CVDs)~\cite{Ward2012HomeMO, Agarwal2011RoleOH}. BP monitoring devices are widely accessible and BP measurement involves obtaining just two numeric values: systolic (SBP) and diastolic (DBP)~\cite{James2014}. Despite its accessibility and importance, BP readings exhibit notable biases due to individual conditions and nonstandard measurement setups~\cite{Mousavi2024-qg}. Many existing BP technologies and standards were established decades ago, among predominantly White and genetically homogeneous populations from the Global North. These clinical standards persist today and are globally applied without adjusting BP readings for individual factors such as biological sex, age, race, height, weight, medical history, or comorbidities like diabetes. Meanwhile, mounting evidence highlights the influence of race and genetics on BP~\cite{Zilbermint2019,Fuchs2011,AnumSaeed2020}. The conventional ranges designating normal and abnormal BP (e.g., below 120/80 mmHg for normal) merely indicate a patient's percentile within the stochastically distributed BP of a population~\cite{James2014}. 
While these standards provide useful guidelines and help identify at-risk patients, they are not fully reliable for diagnosis due to the lack of demographic considerations. As a result, they lack precision as biomarkers for assessing the severity of hypertension or as accurate, quantitative measures of CVD risk.

The scientific literature pertaining to BP is abundant, with thousands of research publications accessible online. These studies collectively present compiled BP data obtained from millions of individuals, encompassing diverse comorbid and demographic factors within various cohorts. Currently,\footnote{Accessed on February 3, 2025.} PubMed alone reports 707,250 instances of the key phrase `\textit{blood pressure}' in its repository of published research. Apparently, not all of these instances contain quantitative BP data. Nevertheless, in clinical cohorts, it is customary to include details such as the number of subjects, inclusion/exclusion criteria, averages, and standard deviations of BPs (as well as other vital signs). Our hypothesis is that the amalgamation of these scientific reports could serve to deduce population-wide distributions of BP across different demographics and individualized factors. Analyzing the published literature on a case-by-case basis would be laborious and infeasible, lack generalizability, and be prone to human error. However, recent advancements in natural language processing (NLP), including large language models (LLMs), offer the opportunity to systematically search and parse the text and reported BP ranges within these publications. This allows for the retrieval of population-wide BP statistics, including the number of subjects, mean and standard deviations of SBP and DBP, as well as information on comorbid factors and demographics. 

In this research, we investigate the viability of employing LLMs for the automated extraction of information related to BP from biomedical literature, with a particular focus on biological sex-based distinctions in BP statistics. Our primary goal is to assess how well a zero-shot LLM (i.e., without any tuning) performs compared to traditional few-shot models in extracting and linking BP measurements (e.g., means and standard deviations) with biological sex distinctions from text. This is a critical step towards large-scale, data-driven analyses of demographic differences in BP distributions. While traditional NLP approaches often rely on low-shot or rule-based systems that require extensive domain-specific tuning, our approach leverages the inherent capabilities of LLMs to generalize across tasks with minimal or no prior training. The advantages of LLMs lie in their scalability and adaptability, which enable the analysis of vast amounts of scientific literature without requiring the development of task-specific models. Our results demonstrate that LLMs outperform traditional low-shot information extraction systems in this context, showcasing their ability to extract BP-related information more accurately and efficiently. By employing LLMs, we provide a scalable framework for analyzing demographic differences in BP and emphasize the broader utility of LLMs in addressing similar challenges in biomedical research. To the best of our knowledge, this is the first study to apply LLMs for extracting quantitative BP-related information from large-scale biomedical literature. This novel application highlights the transformative potential of LLMs in advancing precision medicine and addressing demographic biases in health research.

\section{Statement of Significance}
\subsection{Problem}
Current BP monitoring tools tend to be biased due to insufficient consideration of demographic factors, such as biological sex, leading to potential inaccuracies in understanding and addressing BP-related healthcare disparities. NLP methods can help curate demography-specific BP values from scientific literature.

\subsection{What is already known}
The scientific literature contains extensive studies on BP trends across demographic groups, including differences based on biological sex. Recent advancements in NLP/LLMs enable systematic exploration and extraction of information from these publications, offering a scalable approach to retrieve population-wide BP statistics.

\subsection{What this paper adds}
This study demonstrates the application of NLP methods, specifically LLMs, to extract sex-specific BP statistics from scientific literature automatically. The novelty lies in leveraging a zero-shot LLM for this task, showcasing its potential for efficient information retrieval. Beyond confirming known BP differences between men and women, the study underscores the variability and gaps in reporting standards across literature. These insights highlight the need for standardized reporting in BP studies to enhance data utility in addressing demographic biases. Moreover, the approach provides a blueprint for applying NLP and LLMs to other medical domains, paving the way for more precise, data-driven health interventions.

\section{Related Work}
Since 2023, motivated by the emergence and success of LLMs, numerous studies leveraged them for NLP tasks related to biomedical and healthcare topics. Kung et al.~\cite{TiffanyH.Kung2023} evaluated the ChatGPT LLM (GPT-3.5) on the United States Medical Licensing Exam and demonstrated that the model showed a high level of concordance and insight in its explanations. Chen et al.~\cite{Chen2023} examined the performance of GPT-3.5 on various neurological exam grading scales, where GPT-3.5 demonstrated ability in evaluating neuro-exams using established assessment scales. Similarly, Dehghani et al.~\cite{Dehghani2023} evaluated the performance of GPT-3.5 on a radiology board-style examination, and GPT-3.5 correctly answered 69\% of the questions. These findings indicate that LLMs might possess the capability to support medical education and potentially assist clinical decision-making. Huang et al.~\cite{Huang2023} used GPT-3.5 to extract pertinent information, like a patient’s unique illnesses or adverse effects from free-text notes written by healthcare professionals, which helps automating diagnosis of dental conditions. 
The study conducted by Russe et al.~\cite{Russe2023} utilized LLMs combined with an information retrieval method to extract Arbeitsgemeinschaft Osteosynthesefragen (AO) codes in radiology reports. The results indicated that the models extracted AO codes at a significantly faster rate than humans, with commendable accuracy. These studies collectively demonstrate the potential of employing LLMs for information extraction to advance medical research. In our work, we employ LLMs to extract BP values and related entities based on biological sex from open-access scientific literature. To the best of our knowledge, this is the first work that attempts to extract BP values from literature texts and link them to biological sex automatically, enabling the large-scale analysis of sex-based differences in BP values.

\begin{figure}
    \centering
    \includegraphics[width=1\linewidth]{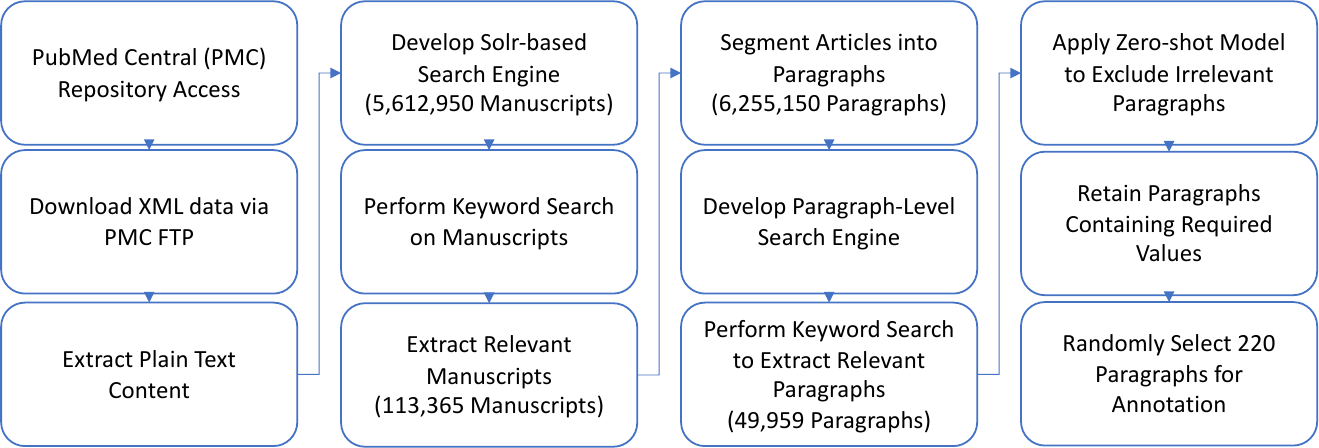}
    \caption{The pipeline for developing a dataset for training and evaluating a system for extracting blood pressure values and biological sex from medical literature text.}
    \label{fig:data_collect}
\end{figure}

\section{Methods}
\subsection{Data Collection}
We created a dataset following the process shown in Figure~\ref{fig:data_collect}. The dataset was sourced from PubMed Central (PMC), an openly accessible repository of biomedical and life sciences journal literature, housed at the U.S. National Institutes of Health's National Library of Medicine (NIH/NLM), accessed via the PMC File Transfer Protocol (FTP) Service.\footnote{ \url{https://ftp.ncbi.nlm.nih.gov/pubmed/baseline/}} The data were stored in Extensible Markup Language (XML) with each publication methodically organized to include the title, authors, digital object identifier (DOI), and plain text of the manuscript. We indexed the plain text content in Solr,\footnote{\url{https://solr.apache.org/}}, an open-source search platform enabling full-text search. We performed keyword-based searches on these manuscripts utilizing SQL to identify studies pertinent to blood pressure.

The data collection was executed through a two-step retrieval process. Initially, we developed a Solr-based search engine encompassing all downloaded articles, totaling 5,612,950 manuscripts, each stored as an individual record within the database. Subsequently, a keyword search was conducted targeting three categories: (i) blood pressure, hypertension, or related keywords; (ii) biological sex indicators such as male and female; and (iii) units of BP measurement such as `mmHg' and `cmHg'. This enabled the extraction of 113,365 articles specifically reporting blood pressure data. Next, each article was segmented into multiple paragraphs using XML tags, and another search engine was indexed for paragraph-level searches. The same keyword criteria were applied, resulting in the retrieval of 6,255,150 paragraphs. This second step was designed to reduce the input text length for the model, enhancing time efficiency. Subsequently, we obtained 49,959 relevant paragraphs, each associated with a corresponding article ID.

Given the extensive number of paragraphs, a subset needed to be sampled for data annotation. Random sampling risked a minimal proportion of pertinent articles; hence, a zero-shot model was employed on these paragraphs to exclude those lacking the entities of interest, resulting in a dataset comprising 3127 paragraphs. From this subset, 220 paragraphs were randomly selected for annotation. The sampling did not take into account the \textit{publication types} of the articles to ensure that the model evaluation generalizes across different types of studies. A graduate student with a medical background and expertise in BP literature manually annotated these samples. To ensure quality, an NLP expert from our team independently double-annotated 50 paragraphs. The inter-annotator agreement (IAA) between the two annotators, measured using Cohen's kappa~\cite{McHugh2012}, was 0.74, which can be considered to be substantial agreement~\cite{Viera2005}. In instances where multiple paragraphs were derived from a single article, preference was given to the ones containing the requisite values post-annotation, with the remainder discarded if none met the criteria. We opted for GPT-3.5 due to its demonstrated high precision in the preliminary analyses of our previous study~\cite{Guo2024}, where it was used to automatically extract BP statistics for males and females from a dataset of 25 million abstracts sourced from PubMed. Note that for this article selection step, the model was not asked to extract any values, but to suggest if an article contained BP-related information or not. Ultimately, this process culminated in the creation of a dataset containing 213 articles including 90 cases that met our criteria. 

\subsection{Information Extraction}
The problem was formulated as an information extraction task involving 10 entities representing the following information: 
\begin{itemize}
    \item number of males and females (`\texttt{N\_male}' and `\texttt{N\_female}');
    \item mean and standard deviation of SBP for males and females (`\texttt{SBP\_in\_male\_mean}', `\texttt{SBP\_in\_male\_std}', `\texttt{SBP\_in\_female\_mean}', and `\texttt{SBP\_in\_female\_std}'); and
    \item mean and standard deviation of DBP for males and females (`\texttt{DBP\_in\_male\_mean}', `\texttt{DBP\_in\_male\_std}', `\texttt{DBP\_in\_female\_mean}', and `\texttt{DBP\_in\_female\_std}').
\end{itemize} 

We employed a few-shot learning model and two LLMs to derive entities from scientific literature, with strict methodological controls to prevent data leakage. The few-shot learning model was trained exclusively on a separate training set and evaluated on an independent test set, ensuring no overlap between training and testing data. The LLMs were tested in a zero-shot setting on the same test set without any model retraining. We applied 5-fold cross-validation, randomly splitting the dataset into non-overlapping training (80\%) and test (20\%) sets for each fold. This approach guarantees that no information from the test sets contaminated the model training process. Performance metrics are reported as mean and standard deviations on the test sets across the five folds.

\subsubsection{Few-Shot Learning Model}
We applied a few-shot learning model that utilizes a data augmentation method in combination with the nearest neighbor classifier (DANN)~\cite{Ge2024}.\footnote{The DANN model is publicly available at \url{https://github.com/Yao-Ge-1218AM/Data-Augmentation-with-Nearest-Neighbor-Classifier}.} The model was adapted from a semantically augmented named entity recognition model~\cite{nie-etal-2020-named}, which included an encoding procedure for augmenting semantic information and a gate module for integrating this augmented information into the backbone model. Specifically, during the encoding procedure, the representation of each token in the input sentence was enhanced with the most semantically similar expressions. The gate module then aggregated these token representations across varying contexts. The model demonstrated superior performance compared to other few-shot learning models across multiple tasks~\cite{Ge2024}.

\begin{figure}
    \centering
    \includegraphics[width=\textwidth]{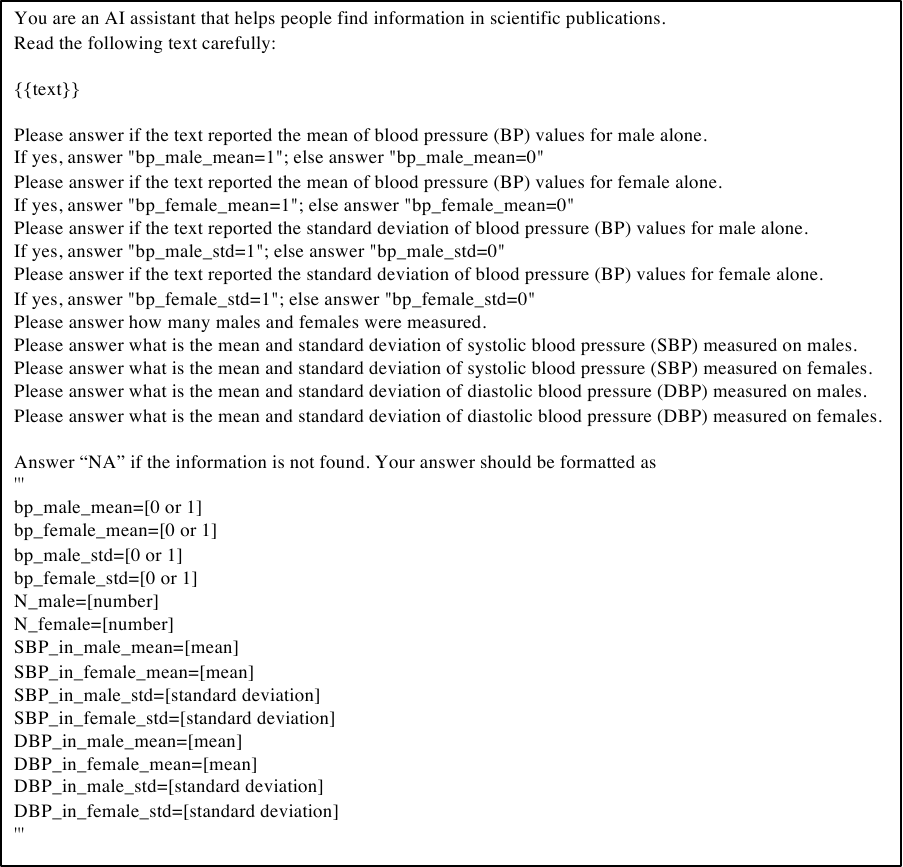}
    \caption{The prompt used for extracting the mean and standard deviation based on biological sex, where `\texttt{\{\{text\}\}}' was the placeholder for the manuscript content.}
    \label{fig:prompt}
\end{figure}

\subsubsection{Large Language Models}
We selected LLaMA3~\cite{llama3modelcard} and GPT-3.5~\cite{Brown2020} for information extraction in zero-shot setting. LLaMA3 is an open-source LLM, pretrained on over 15 trillion tokens of data from publicly available sources. The pretraining data also includes publicly available instruction datasets, as well as over 10 million human-annotated examples. For this study, we selected the LLaMA3 model variant with 70 billion parameters based on our available computational resources. GPT-3.5, in contrast, is a closed-source large language model, designed to understand and generate human-like text. It can effectively perform various NLP tasks such as answering questions, writing content, and assisting with coding. It powers ChatGPT\footnote{Released on November 30, 2022}, a popular AI chatbot. 

The input for LLM, referred to as the \texttt{prompt}, consisted of a plain text description providing specific instructions for the model. The prompt included explicit guidelines for organizing the answer in a particular format to facilitate post-generation parsing and retrieval of entity values. We conducted prompt engineering to refine the prompt. As discussed in our preliminary work~\cite{Guo2024}, LLMs tend to generate values for all requested entities, even if these values are not explicitly mentioned in the input. This is known as \textit{hallucination}, which is a known issue with existing LLMs. To mitigate such errors, we revised the prompt to instruct the model to explicitly state whether the text mentioned the entity or not. The corresponding flags in the answer were `\texttt{bp\_male\_mean}`, `\texttt{bp\_female\_mean}`, `\texttt{bp\_male\_std}`, and `\texttt{bp\_female\_std}`. We only used the other predicted entities if the model predicted all four of these flags as \texttt{True}; otherwise, the other entities were considered as \texttt{False}. The final details of the prompt are provided in Figure \ref{fig:prompt}. After generating answers for each article, a simple parsing script using regular expressions was executed to extract the values for each entity.

Additionally, in our preliminary study, the LLM often performed its own calculations based on values present in the scientific literature and presented these calculated values. For example, we observed that when some articles contained multiple blood pressure values for males or females, the model attempted to automatically compute the average. To address this issue, we performed post-processing on the values of all entities. If any of the predicted values did not appear in the text, all entities were set to zero.

\begin{table}[tb]
    \centering
    \caption{Precision, recall, and F$_1$ scores for different models on the system level.}
    \begin{tabular}{|l|c|c|c|}
        \hline
        \textbf{Model}            & \textbf{Precision}       & \textbf{Recall}          & \textbf{F$_1$}             \\ \hline
        DANN  & 0.34 ($\pm$0.01)    & 0.28 ($\pm$0.01)    & 0.30 ($\pm$0.01)    \\ \hline
        LLaMA3       & 0.87 ($\pm$0.00)    & 0.84 ($\pm$0.00)    & 0.85 ($\pm$0.00)    \\ \hline
        GPT-3.5           & 0.86 ($\pm$0.02)    & 0.55 ($\pm$0.04)    & 0.67 ($\pm$0.04)    \\ \hline
    \end{tabular}
    \label{tab:model_performance}
\end{table}

\begin{table}[htpb!]
  \centering
  \caption{Precision, recall, and F$_1$ scores for different models on the entity level.}
  \label{tab:ent_performance_metrics}
  \begin{tabular}{|l|c|c|c|}
    \hline
    \textbf{Entity} & \textbf{Precision} & \textbf{Recall} & \textbf{F$_1$} \\
    \hline
    \multicolumn{4}{|c|}{\textbf{DANN}} \\
    \hline
    \texttt{N\_male}             & 0.05 ($\pm$0.01) & 0.38 ($\pm$0.04) & 0.09 ($\pm$0.01) \\
    \texttt{N\_female}           & 0.05 ($\pm$0.01) & 0.27 ($\pm$0.05) & 0.08 ($\pm$0.01) \\
    \texttt{SBP\_in\_male\_mean} & 0.04 ($\pm$0.01) & 0.25 ($\pm$0.08) & 0.07 ($\pm$0.01) \\
    \texttt{SBP\_in\_female\_mean} & 0.05 ($\pm$0.01) & 0.26 ($\pm$0.07) & 0.07 ($\pm$0.01) \\
    \texttt{SBP\_in\_male\_std}  & 0.07 ($\pm$0.01) & 0.41 ($\pm$0.04) & 0.11 ($\pm$0.01) \\
    \texttt{SBP\_in\_female\_std} & 0.08 ($\pm$0.01) & 0.48 ($\pm$0.04) & 0.13 ($\pm$0.01) \\
    \texttt{DBP\_in\_male\_mean} & 0.09 ($\pm$0.01) & 0.41 ($\pm$0.05) & 0.15 ($\pm$0.02) \\
    \texttt{DBP\_in\_female\_mean} & 0.08 ($\pm$0.01) & 0.38 ($\pm$0.05) & 0.14 ($\pm$0.01) \\
    
    \texttt{DBP\_in\_male\_std}  & 0.05 ($\pm$0.01) & 0.22 ($\pm$0.05) & 0.09 ($\pm$0.02) \\
    \texttt{DBP\_in\_female\_std} & 0.07 ($\pm$0.01) & 0.19 ($\pm$0.05) & 0.10 ($\pm$0.02) \\
    \hline
    \multicolumn{4}{|c|}{\textbf{LLaMA3}} \\
    \hline
    \texttt{N\_male}             & 0.90 ($\pm$0.00) & 0.93 ($\pm$0.00) & 0.92 ($\pm$0.00) \\
    \texttt{N\_female}           & 0.90 ($\pm$0.00) & 0.94 ($\pm$0.00) & 0.92 ($\pm$0.00) \\
    \texttt{SBP\_in\_male\_mean} & 0.91 ($\pm$0.00) & 0.94 ($\pm$0.02) & 0.92 ($\pm$0.01) \\
    \texttt{SBP\_in\_female\_mean} & 0.88 ($\pm$0.01) & 0.88 ($\pm$0.03) & 0.88 ($\pm$0.01) \\
    
    \texttt{SBP\_in\_male\_std}  & 0.92 ($\pm$0.00) & 0.90 ($\pm$0.01) & 0.91 ($\pm$0.00) \\
    \texttt{SBP\_in\_female\_std} & 0.94 ($\pm$0.00) & 0.88 ($\pm$0.02) & 0.91 ($\pm$0.01) \\
    \texttt{DBP\_in\_male\_mean} & 0.93 ($\pm$0.00) & 0.96 ($\pm$0.00) & 0.95 ($\pm$0.00) \\
    \texttt{DBP\_in\_female\_mean} & 0.91 ($\pm$0.00) & 0.92 ($\pm$0.00) & 0.92 ($\pm$0.00) \\
    \texttt{DBP\_in\_male\_std}  & 0.92 ($\pm$0.00) & 0.92 ($\pm$0.00) & 0.92 ($\pm$0.00) \\
    \texttt{DBP\_in\_female\_std} & 0.92 ($\pm$0.00) & 0.95 ($\pm$0.01) & 0.93 ($\pm$0.01) \\
    \hline
    \multicolumn{4}{|c|}{\textbf{GPT-3.5}} \\
    \hline
    \texttt{N\_male}             & 0.88 ($\pm$0.07) & 0.58 ($\pm$0.01) & 0.69 ($\pm$0.01) \\
    \texttt{N\_female}           & 0.89 ($\pm$0.05) & 0.68 ($\pm$0.01) & 0.77 ($\pm$0.00) \\
    \texttt{SBP\_in\_male\_mean} & 0.85 ($\pm$0.07) & 0.59 ($\pm$0.01) & 0.69 ($\pm$0.01) \\
    \texttt{SBP\_in\_female\_mean} & 0.85 ($\pm$0.06) & 0.55 ($\pm$0.01) & 0.66 ($\pm$0.00) \\
    \texttt{SBP\_in\_male\_std}  & 0.91 ($\pm$0.05) & 0.61 ($\pm$0.02) & 0.73 ($\pm$0.01) \\
    \texttt{SBP\_in\_female\_std} & 0.93 ($\pm$0.04) & 0.60 ($\pm$0.01) & 0.73 ($\pm$0.00) \\
    \texttt{DBP\_in\_male\_mean} & 0.88 ($\pm$0.07) & 0.64 ($\pm$0.01) & 0.74 ($\pm$0.01) \\
    \texttt{DBP\_in\_female\_mean} & 0.87 ($\pm$0.06) & 0.52 ($\pm$0.01) & 0.65 ($\pm$0.01) \\
    
    \texttt{DBP\_in\_male\_std}  & 0.88 ($\pm$0.07) & 0.49 ($\pm$0.01) & 0.62 ($\pm$0.00) \\
    \texttt{DBP\_in\_female\_std} & 0.91 ($\pm$0.05) & 0.55 ($\pm$0.06) & 0.68 ($\pm$0.05) \\
    \hline
  \end{tabular}
\end{table}

\subsubsection{Evaluation Metrics}
Final model performance was assessed on the test set using the precision, recall and F$_1$ score metrics at both the entity level and system level. 
Precision measures the accuracy of the extracted information. It is calculated as the number of correct (relevant and correctly identified) pieces of information extracted divided by the total number of pieces of information extracted. Higher precision indicates fewer false positives. Recall assesses the system’s ability to identify all relevant information. It is calculated as the number of correct pieces of information extracted divided by the total number of relevant pieces of information in the text. Higher recall indicates fewer false negatives. F$_1$ score is the harmonic mean of precision and recall. It provides a single metric that balances both precision and recall, giving an overall sense of the system's performance. These metrics provided a standardized way to compare the performance of different information extraction systems used in the Message Understanding Conference (MUC) series, particularly MUC-5~\cite{Chinchor1993}, assessing both the accuracy and completeness of the extracted information. For each model, we ran the experiment five times with different random initialization and reported the mean and standard deviation of the metrics across these runs.

\subsection{BP Variation Analysis}
 In a recent research study on bias in BP measurement technologies, authors collected information by manually investigating just 20 papers and analyzing BP data related to biological sex~\cite{Mousavi2024-qg}. They represented the distribution of BP information using ellipses, with each ellipse associated with a specific study. The centers of the ellipses correspond to the mean of DBP and SBP, while the horizontal and vertical radii correspond to the standard deviations of SBP and DBP, respectively. Subsequently, they used heatmaps to visualize distributions. They also employed Gaussian mixture models (GMM), a statistical method used for clustering and density estimation, to explore the data further, assuming that the reported BP values follow a normal distribution. In this study, we employed GMM to generate contour plots for representing SBP and DBP associated with males and females, automatically extracted from the scientific literature. Our hypothesis is that since BP is influenced by factors such as age, sex, race, pregnancy, lifestyle, and socioeconomic status, a one-size-fits-all threshold for normal and abnormal BP may not be optimal for all populations. The contour plots represent the 95\% percentile range of the population that fall within each contour, which is generally an indicator of data scatter around the reported mean values.

\section{Results}
Tables \ref{tab:model_performance} and \ref{tab:ent_performance_metrics} show the performance of three models at the system level and on the entity level. For DANN, the model shows low performance across all entities, with the highest F$_1$ score (0.14) for the mean DBP in females. LLaMA3 is significantly better than the few-shot learning method across all entities. The recall and precision values are notably higher, leading to better F$_1$ scores. GPT-3.5 achieves moderate performance, better than DANN but generally lower than LLaMA3. We also noticed that GPT-3.5 achieved comparably high precision but significantly lower recall compared to LLaMA3. Additionally, we observed that the standard deviation of the evaluation metrics for LLaMA3 are lower than DANN and GPT-3.5. In summary, LLaMA3 outperforms both DANN and GPT-3.5 across all entities, especially in terms of precision and recall. GPT-3.5 shows moderate performance, generally better than DANN but not as good as LLaMA3. DANN has the lowest performance across all entities.

\begin{figure}[tb]
    \centering
    \includegraphics[width=\textwidth]{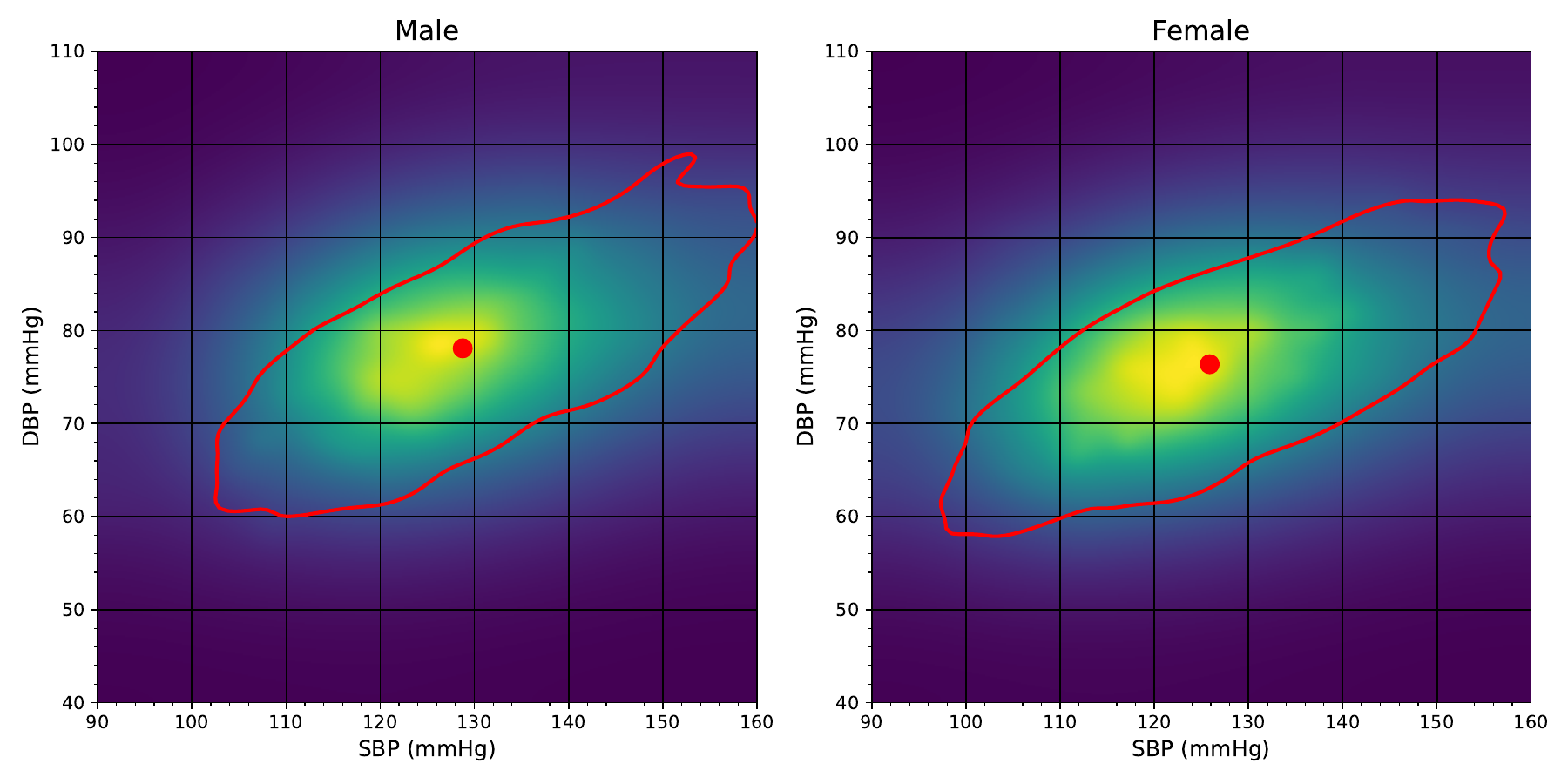}
    \caption{Comparisons of blood pressure distributions between sexes are presented through heatmaps and contour plots, employing GMM fitted on the BP values retrieved by the proposed LLM-based technique. The contour plots represent the 95\% percentile range of the population that fall within each contour.}
    \label{fig: heatmapandcontourplot}
\end{figure}

Figure~\ref{fig: heatmapandcontourplot} illustrates heatmaps and contour plots associated with BP data for each sex category, facilitating a comparison of BP results between males and females. The contour lines on these plots highlight areas where BP values are more concentrated. As depicted in the figure, according to the peak points on the heatmaps, it shows that males tend to exhibit higher BP values than females. 

\section{Discussion}
This study presents a comprehensive comparison of the performance of three information extraction models---a few-shot learning method (DANN), GPT-3.5, and LLaMA3---for multiple entities related to BP based on biological sex from scientific literature.
The few-shot learning method, which does not include a generative LLM, despite its potential in scenarios with limited training data, demonstrates the lowest performance among the three models across all entities. The highest recall achieved by this model is 0.41 for the mean DBP in males, and the highest precision is 0.09 for the same metric. However, the F$_1$ score, which balances precision and recall, remains low at 0.15. This indicates that the model struggles to provide accurate and reliable predictions even when it occasionally identifies relevant instances.
LLaMA3 significantly outperforms the few-shot learning method across all evaluated entities. The model's high recall and precision values result in superior F$_1$ scores, suggesting a robust performance in extracting the BP related entities from the scientific literature. Notably, LLaMA3 exhibits the highest F$_1$ score of 0.95 for the standard deviation of DBP in males. The standard deviation of the evaluation metrics being 0 for this model further underscores its consistency and reliability. GPT-3.5 shows moderate performance, standing between the few-shot learning method and LLaMA3. It generally performs better than DANN but not as well as LLaMA3 especially for recall. 

LLaMA3's superior performance across most entities, particularly in precision and recall, positions it as the most effective model in this study. Its consistency, as indicated by the standard deviation of 0 in evaluation metrics, further demonstrates its robustness and reliability for practical applications. In contrast, GPT-3.5, with its moderate performance, demonstrates that it can serve as a viable middle-ground solution, particularly in contexts where LLaMA3's higher computational demands might be a limiting factor. In summary, the LLaMA3 model outperforms both the few-shot learning method (DANN) and GPT-3.5. Our findings indicate the potential to develop a highly accurate information extraction system for analyzing BP values from scientific literature using LLMs. Importantly, this excellent performance was achieved in a zero-shot setting, demonstrating the superiority of LLMs compared to more traditional low-shot information extraction systems. It is particularly noteworthy that LLaMA3's excellent performance was achieved in a zero-shot setting, without task-specific fine-tuning. This finding is significant as it demonstrates that large language models can now outperform specialized few-shot learning approaches like DANN in domain-specific information extraction tasks. Although DANN represents a strong baseline among few-shot learning methods as established in previous benchmarks~\cite{Ge2024}, the substantial performance gap observed suggests a fundamental advantage of large-scale pre-trained models for complex information extraction tasks in specialized domains.

The performance difference between GPT-3.5 and LLaMA3 may be influenced by the prompt design, as it was specifically optimized for LLaMA3. Prompt optimization for this model particularly improved recall, as shown in Table \ref{tab:model_performance}. LLaMA3 also demonstrated more robust performance across folds, leading to substantially lower standard deviations particularly for precision and F$_1$ score, compared to GPT-3.5. It is likely that elaborate prompt engineering will improve the performance of GPT-3.5 for this task. However, the excellent performance of LLaMA3 and its open-source license makes it ideal for our objective of extracting BP values from big scientific literature data. In future work, we plan to explore prompt optimization specifically targeting the F$_1$ score to better understand its impact on model performance across different models.

The contour plots show that males tend to exhibit higher BP values than females, aligning with consistent findings from numerous previous studies~\cite{Sandberg2012,Reckelhoff2001-td,Mousavi2024-qg}. It suggests that by analyzing BP distributions across diverse demographic subgroups, we can identify variations that may be overlooked in generalized clinical guidelines. This work serves toward this broader aim and strengthens our hypothesis by allowing us to establish more personalized BP reference ranges, which can improve the accuracy of hypertension diagnosis and risk stratification. Additionally, understanding these demographic variations helps contextualize observed BP differences and reduces the risk of misclassification, ultimately leading to better-targeted interventions and healthcare strategies.

\section{Limitations}

This study has several important limitations that should be considered when interpreting our findings.
First, our investigation focused exclusively on biological sex as a factor in blood pressure (BP) reporting, without addressing other clinically significant variables. We acknowledge that BP, as a vital sign rather than a demographic variable, is influenced by multiple factors including age, race/ethnicity, lifestyle, diet, socioeconomic status, and geographic location. The complex interrelationships between these variables mean that examining sex differences in isolation provides an incomplete picture of BP distributions in populations. A more comprehensive approach would consider these confounding factors to meaningfully interpret sex-based differences in BP.
Second, our extraction focused only on mean and standard deviation measures of BP, overlooking other important statistical descriptors such as median, interquartile range (IQR), skewness (third moment), and kurtosis (fourth moment). These additional measures provide crucial information about the shape and characteristics of BP distributions that may not be captured by means and standard deviations alone.
Third, while DANN has demonstrated strong performance relative to other few-shot learning approaches in previous research, our experimental design included only one representative few-shot learning model. Although this choice was justified based on DANN's established capabilities in similar extraction tasks, the comparison between LLMs and traditional few-shot methods would be strengthened by evaluating against multiple few-shot learning architectures.
Fourth, our evaluation used a curated dataset from scientific literature in a specific domain, which may not fully represent the challenges of extracting information from other types of clinical or scientific documents with varying structures, terminologies, and reporting conventions. Also, there can be redundancy in the publication due to same dataset and bias, which we do not have a mechanism to handle at the moment. In the future, we will explore how to remove redundancy in the data.  
We faced practical constraints related to computational resources when conducting experiments with parameter-dense models like LLaMA3-70B, which limited our ability to expand the scope of this study. Despite these limitations, our findings provide valuable insights into the comparative effectiveness of different model architectures for specialized information extraction tasks. We leave the extension to additional physiological and contextual factors, broader statistical measures, diverse few-shot learning approaches, and testing across different document types as important directions for future work.

\section{Reproducibility and Code Availability}
To facilitate reproducibility and encourage further research, we have made the full source code, along with detailed documentation of the experimental setup, available in a public GitHub repository: \url{https://github.com/yguo0102/blood_pressure_project}. The repository includes all scripts used for data preprocessing, model running, post-processing, and evaluation, as well as specifications of the computational environment. We have also released the annotated examples in the repository to support result reproduction. This ensures that all methodological steps described in the manuscript can be independently validated and extended.  We believe that sharing our implementation not only strengthens the transparency of our work but also supports collaborative development and application in related research areas.

\section{Conclusion}
In this research, we investigated the possibility of applying LLMs in a zero-shot setting to extract the mean and standard deviation of BP values with a biological sex-based distinction. Also, we compared a few-shot learning method and two LLMs, and the results demonstrate the superiority of LLMs compared to more traditional low-shot information extraction systems. Furthermore, we conducted variation analysis on the extracted information. The results suggest that males might have higher BP levels than females, which is in line with population-wide studies using actual BP measurements from electronic health records~\cite{mousavi2024learning}. Importantly, this study showed that it is possible to leverage LLMs to analyze a substantial volume of scientific literature to extract BP-related information based on various demographic factors. Future work will expand this study to investigate the BP variations based on other demographic factors using the scientific literature. A similar methodology can be explored for retrieving other vital measurements from the scientific literature.

%
%
\bibliographystyle{elsarticle-harv} 
\bibliography{mybibliography}

\newpage
\appendix
\section{LLM Examples}
\begin{figure}
    \centering
    \includegraphics[width=\textwidth]{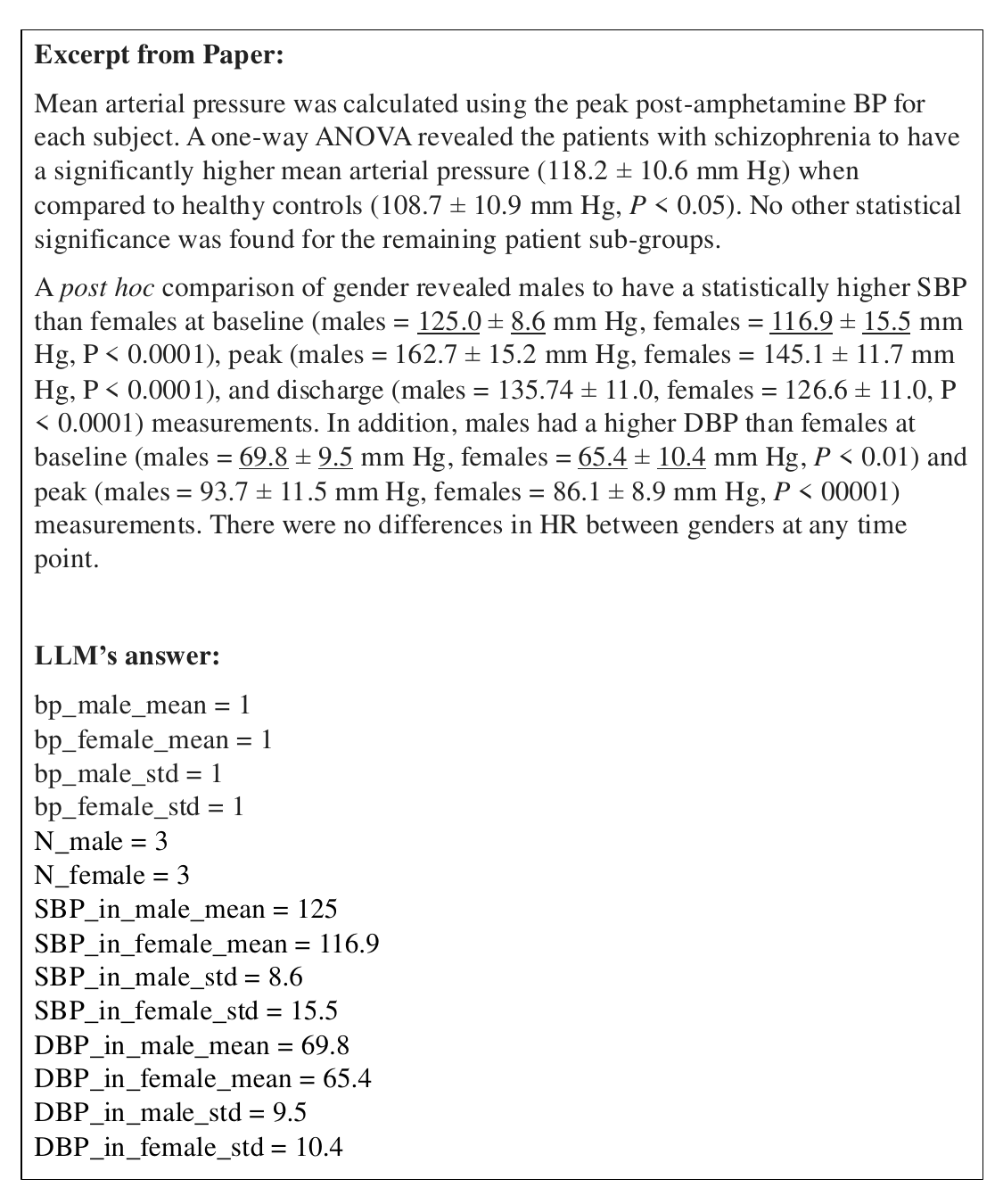}
    \caption{From this study, the LLM was able to extract the appropriate mean and standard deviation values based on biological sex within the text of the research paper. The example is adapted from Weidner et al. \cite{Weidner2015-zn}}.
\end{figure}

\begin{figure}
    \centering
    \includegraphics[width=\textwidth]{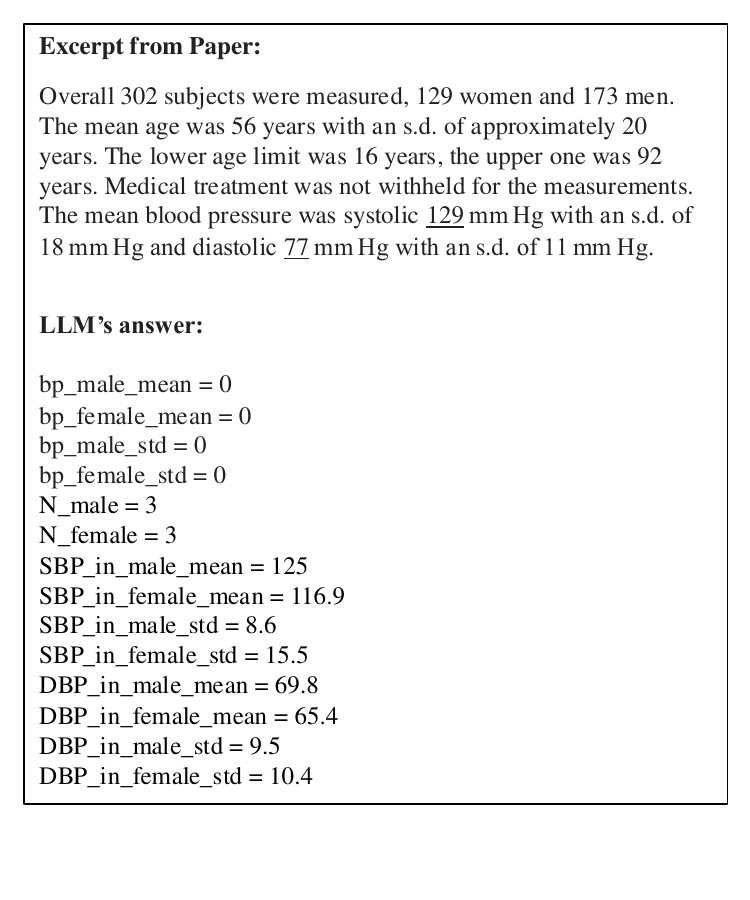}
    \caption{From this study, the LLM was not able to extract the appropriate mean and standard deviation values based on biological sex within the text of the research paper. The example is adapted from Wassertheurer et al. \cite{Wassertheurer2010}}.
\end{figure}

\end{document}